\documentclass{article}

\usepackage{genai4health_2026}

\usepackage[utf8]{inputenc}
\usepackage[T1]{fontenc}
\usepackage{hyperref}
\usepackage{url}
\usepackage{booktabs}
\usepackage{amsmath,amsfonts,amssymb}
\usepackage{nicefrac}
\usepackage{microtype}
\usepackage{xcolor}
\usepackage{graphicx}
\usepackage{multirow}
\usepackage{array}
\usepackage{enumitem}
\usepackage{caption}
\usepackage{subcaption}
\usepackage{mathtools}
\usepackage{pifont}
\hypersetup{colorlinks=true,citecolor=black,linkcolor=black,urlcolor=blue}

\title{From Scores to Safe Sets:\\Exact Safety-Constrained Decoding for Medication Recommendation}
\author{%
\normalfont
\href{https://openreview.net/profile?id=~Animesh_Agarwal2}{Animesh Agarwal}
\quad
\href{https://openreview.net/profile?id=~Meysam_Ghaffari1}{Meysam Ghaffari}
\quad
\href{https://openreview.net/profile?id=~Nina_Fatehi1}{Nina Fatehi}\\
\href{https://openreview.net/profile?id=~Anushka_Swarup1}{Anushka Swarup}
\quad
\href{https://openreview.net/profile?id=~Carlos_Morato1}{Carlos Morato}\thanks{Correspondence to: \href{mailto:carlos.morato@optum.com}{\texttt{carlos.morato@optum.com}}}\\[2pt]
Optum
}

\newcommand{\pcls}{\ensuremath{\pi_{\mathrm{cls}}}}

\newcommand{\cmark}{\ding{51}}

\begin{document}
\maketitle

\begin{abstract}
Recent generative medication recommendation systems ultimately produce sets of drugs, but drug–drug interaction (DDI) constraints depend on the set as a whole. A natural post-hoc approach is to first predict medications and then locally repair interacting pairs by removing one drug at a time. We ask whether this local repair step itself introduces avoidable error. Using fixed medication scores from FLAME's frozen classifier on MIMIC-III, we compare increasingly strong local repair strategies with an exact global decoder while holding the scores, output size, and DDI constraint fixed. The global decoder jointly selects the highest-scoring feasible medication set using mixed-integer linear programming. Across validation and test splits, global decoding consistently achieves higher Jaccard similarity than greedy repair, degree-aware repair, and degree-aware repair strengthened with exhaustive 1-swap local search. Its advantage is small under loose DDI constraints and becomes larger as the constraint tightens; on the test split it reaches +0.0064 Jaccard over the strongest local baseline at the strict DDI operating point. The finding also holds across a fine accuracy--DDI sweep at matched realized DDI exposure and under three score representations. At the strict setting, global decoding retains more true-positive medications while removing more false positives. These results show that medication recommendation is not only a representation-learning problem: even with a frozen clinical model, how its scores are converted into a constrained medication set can materially affect the final prediction.
\end{abstract}

\section{Introduction}

Medication recommendation from electronic health records (EHRs) is a canonical set-valued prediction problem. Given diagnoses, procedures, and longitudinal context, a model must recommend a combination of medications rather than a single label. This distinction matters because interactions are relational, duplicate or incompatible therapies arise at the set level, and the predicted medication set may need to satisfy multiple constraints simultaneously. A large literature has therefore developed increasingly structured medication recommenders, including sequence models, memory networks, graph-augmented models, molecular encoders, set-to-set predictors, causal representations, multimodal models, and more recently large-language-model (LLM) systems \citep{zhang2017leap,le2018dmnc,shang2019gamenet,shang2019gbert,yang2021micron,yang2021safedrug,wu2022cognet,tan2022foursdrug,wu2023dapsnet,yang2023molerec,chen2023carmen,zhao2024raremed,kim2024vita,li2024causalmed,tan2024nla,liu2024leader,li2025timeaware,wu2025armr,zhao2025lamo,fan2025flame,lv2026giant,huh2026pace}. A recurring objective across this literature is to preserve recommendation accuracy while reducing harmful drug--drug interactions (DDIs) \citep{shang2019gamenet,yang2021safedrug,tan2022foursdrug,wu2023dapsnet,yang2023molerec,chen2023carmen}.

Throughout this paper, references to ``safety'' refer only to the benchmark DDI constraint used in the evaluation. They should not be interpreted as establishing comprehensive clinical safety, clinical utility, or a prescribing recommendation for the resulting medication sets.

Most prior work attacks this tradeoff primarily through \emph{learning}: DDI graphs are incorporated into representation learning, molecular structure is encoded to improve drug representations, DDI penalties are added to the training objective, or list-wise objectives are used to align recommendation behavior \citep{shang2019gamenet,yang2021safedrug,yang2023molerec,chen2023carmen,fan2025flame}. A systematic survey of medication recommendation also highlights a basic difficulty with DDI-based post-processing: removing one interacting drug can discard treatment value elsewhere in the regimen \citep{ali2023survey}. Prior work therefore mainly asks how to learn better scores or safer representations. We ask a complementary question: \emph{once a patient-specific scorer already exists, does error arise when those scores are converted into a DDI-constrained medication set?} Our central finding is that it does: even strong local repair leaves measurable accuracy on the table relative to global constrained decoding. For modern generative medication recommenders, this final decoding step is the point where model-derived scores are converted into a coupled medication set.

A natural local strategy is to repair conflicts after prediction. Start from the model's preferred set, identify a DDI conflict, remove one of the two drugs, and repeat until a target DDI level is reached. A stronger heuristic considers each drug's conflict degree and removes high-degree nodes first, and stronger still, a local-search procedure repairs bad greedy commitments after the fact by searching every feasible swap. All three remain local: a removal is committed before the algorithm knows how that choice will constrain later removals and replacements.

The problem becomes combinatorial when conflicts are dense. In our evaluation regime, output sets contain approximately 27 medications on average, and a typical predicted set contains dozens of simultaneously active DDI edges. Removing one drug may resolve many edges at once but may also force the loss of a highly supported medication; retaining a high-scoring ``hub'' drug can make several other high-scoring medications mutually infeasible. The quality of a local decision therefore depends on the entire conflict graph, which motivates a global formulation that selects the full set jointly.

\begin{figure*}[t]
    \centering
    \includegraphics[width=\textwidth]{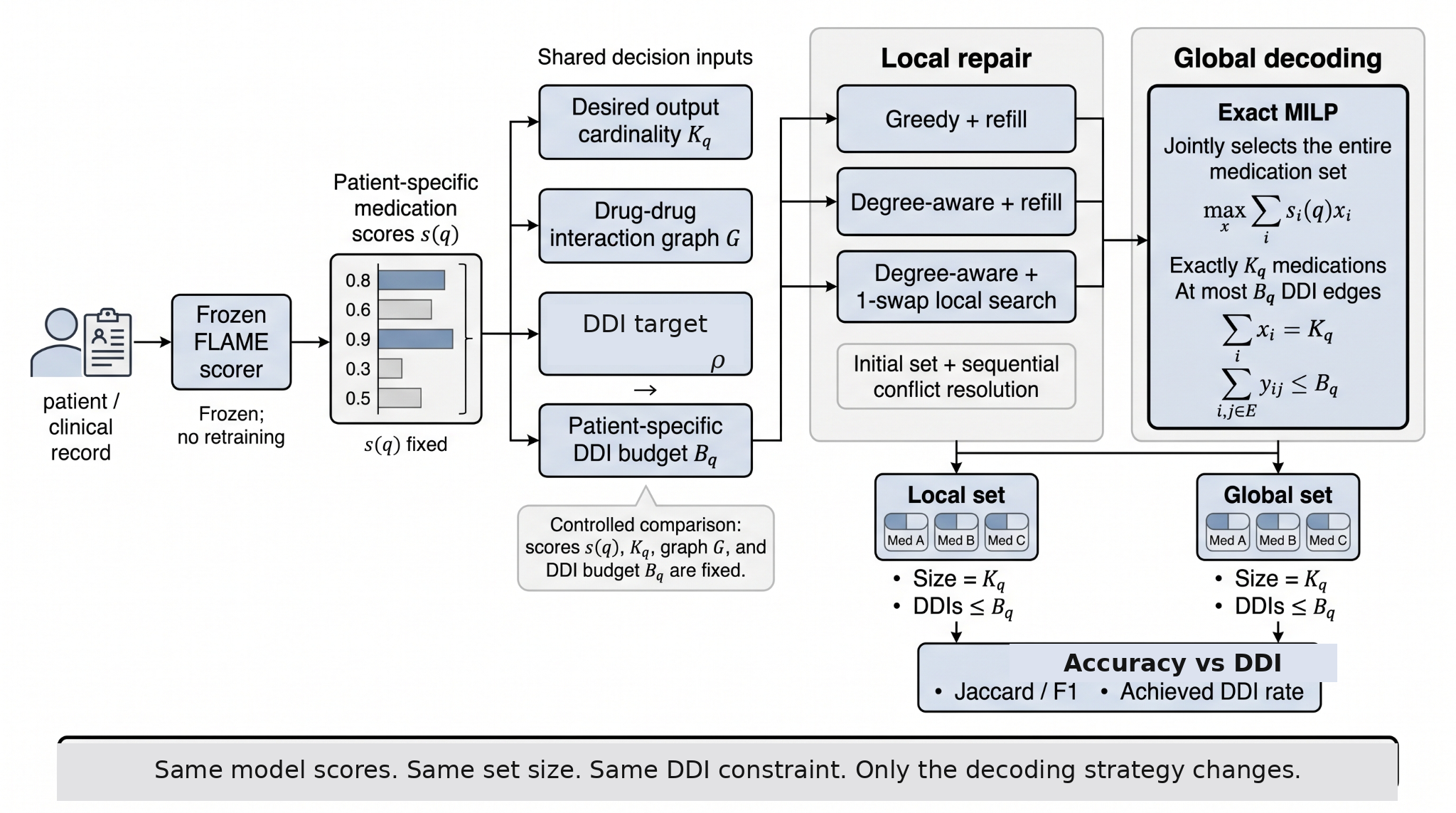}
    \caption{
    \textbf{Controlled comparison of local repair and global DDI-constrained decoding.}
    A frozen FLAME scorer produces patient-specific medication scores $s(q)$. 
    All decoding methods use the same scores, desired output cardinality $K_q$, drug--drug interaction (DDI) graph $G$, and patient-specific DDI budget $B_q$. 
    Local methods construct or repair the medication set through sequential decisions, whereas the exact MILP jointly selects the entire set to maximize the frozen model score while satisfying the same cardinality and DDI constraints. 
    The resulting medication sets are compared using set accuracy and achieved DDI rate. 
    Thus, the experiment isolates the effect of the decoding strategy while keeping the underlying model scores and DDI constraints fixed.
    }
    \label{fig:decoding_overview}
\end{figure*}

We formulate DDI-constrained decoding as a mixed-integer linear program (MILP). For each patient visit $q$, a frozen scorer provides one score $s_i(q)$ per medication. We select exactly $K_q$ medications to maximize the total score, subject to an explicit budget $B_q$ on selected DDI edges. Every decoder returns the same number of medications and the same DDI budget, so the comparison isolates one factor: local versus global selection. Figure~\ref{fig:decoding_overview} summarizes this shared setup.

We instantiate the scorer with FLAME's released drug-level classifier, \pcls{} \citep{fan2025flame}, on MIMIC-III \citep{johnson2016mimic}, using the TWOSIDES-derived DDI graph released with FLAME's preprocessing \citep{fan2025flame,tatonetti2012ddi}. All methodological choices are made on a training split; a disjoint validation split confirms the phenomenon before a held-out test split is scored once. The three splits follow a 4:1:1 train/validation/test ratio over approximately 14,000 total visits. We compare exact global decoding with (i) greedy repair, (ii) degree-aware repair, and (iii) degree-aware repair strengthened with exhaustive 1-swap local search. Across loose, moderate, and strict DDI budgets, the ordering $J_{\mathrm{global}} > J_{\mathrm{1\text{-}swap}} > J_{\mathrm{degree}} > J_{\mathrm{greedy}}$ holds on both validation and test. The advantage grows as the DDI constraint tightens, holds when methods are compared at matched realized (not merely target) DDI exposure across a fine budget sweep, and holds under three different score representations.

Our contributions are:
\begin{enumerate}[leftmargin=*,itemsep=0pt,topsep=2pt,parsep=0pt,partopsep=0pt]
    \item \textbf{A decoding-level view of DDI-constrained recommendation.} We hold the patient--drug scorer, output cardinality, and DDI budget fixed and isolate the final set-selection problem.
    \item \textbf{A controlled comparison of local and global decoding.} We compare exact MILP decoding with greedy repair, degree-aware repair, and degree-aware repair strengthened by exhaustive 1-swap local search.
    \item \textbf{Consistent and distinguishable evidence across splits and operating points.} The advantage reproduces on validation and test, persists at matched realized DDI exposure over a fine DDI sweep, and holds under three score parameterizations.
    \item \textbf{A mechanistic account.} Under strict DDI-constraint pressure, global decoding retains more true positives while removing more false positives, consistent with the cost of locally committing to coupled medication choices.
\end{enumerate}

\section{Related work}

\paragraph{Medication recommendation from EHRs.}
Medication recommendation has progressed from instance- and sequence-based prediction to increasingly structured set prediction. LEAP framed treatment-combination recommendation as sequential generation under multimorbidity \citep{zhang2017leap}; DMNC modeled asynchronous longitudinal views with external memory \citep{le2018dmnc}; GAMENet fused longitudinal patient representations with EHR and DDI graphs \citep{shang2019gamenet}; and G-BERT introduced graph-augmented pretraining for medical-code representations \citep{shang2019gbert}. MICRON focused on medication changes across visits \citep{yang2021micron}, while COGNet used conditional generation with copy-or-predict behavior over historical medications \citep{wu2022cognet}. DAPSNet added patient-similarity retrieval to longitudinal representation learning \citep{wu2023dapsnet}. More recent work improves visit selection, rare-disease performance, causal patient-state modeling, or temporal treatment reasoning, including VITA, RAREMed, CausalMed, and time-aware dynamic-treatment approaches \citep{kim2024vita,zhao2024raremed,li2024causalmed,li2025timeaware}. Surveys summarize the rapid expansion of architectures and evaluation practices in this area \citep{ali2023survey}.

\paragraph{DDI-aware and set-structured recommendation.}
DDI-aware recommendation is often incorporated during training. GAMENet integrates a DDI graph into its memory architecture \citep{shang2019gamenet}. SafeDrug uses molecular graph encoders together with a controllable DDI loss \citep{yang2021safedrug}. 4SDrug explicitly targets small and safe drug sets \citep{tan2022foursdrug}. MoleRec learns substructure-aware molecular representations and uses an annealed objective to balance recommendation accuracy and DDI safety \citep{yang2023molerec}. Carmen designs context-aware molecular and DDI graph embeddings \citep{chen2023carmen}, and DAPSNet includes a DDI-guided loss in a longitudinal patient-similarity model \citep{wu2023dapsnet}. These methods show that DDI information can improve the learned predictor. Our focus is orthogonal: we freeze the predictor and study the optimization problem induced at the final set-selection stage.

\paragraph{Language models and generative medication recommendation.}
Recent systems bring pretrained language models and LLMs into medication recommendation. LEADER distills knowledge from an LLM into a medication recommender \citep{liu2024leader}; NLA-MMR aligns natural-language patient and drug representations with molecular information \citep{tan2024nla}; LAMO fine-tunes LLMs for medication recommendation \citep{zhao2025lamo}; FLAME adds list-wise alignment and step-wise GRPO to a point-wise LLM-based medication classifier \citep{fan2025flame}; GiantMed uses deep-model guidance to activate LLM reasoning near recommendation boundaries \citep{lv2026giant}; and PACE-RAG studies patient-aware retrieval for clinical drug recommendation \citep{huh2026pace}. Our experiments use FLAME only as a source of fixed patient--drug scores. The proposed decoder neither trains nor queries an additional language model, making it a modular inference layer that can in principle sit on top of any scorer.

\paragraph{Constrained inference.}
Mixed-integer optimization is a standard tool for exact discrete decision making under linear constraints \citep{nemhauser1988integer}. The medication-recommendation literature frequently describes the task itself as combinatorial \citep{zhang2017leap,tan2022foursdrug,yang2023molerec,fan2025flame}, but the dominant safety interventions in learned systems act through training losses, graph representations, sequential generation, or heuristic post-processing. We isolate the final inference problem and solve the fixed-score, fixed-cardinality, DDI-budgeted set-selection objective exactly, permitting a controlled comparison in which the score function is identical and only the decoder changes.

\section{Problem formulation}

Let $\mathcal{M}=\{1,\ldots,M\}$ denote the medication vocabulary, with $M=151$ in our experiments. For each patient visit $q$, a frozen model provides scores
\begin{equation}
    s(q) = [s_1(q),\ldots,s_M(q)] \in \mathbb{R}^{M},
\end{equation}
where larger $s_i(q)$ indicates stronger model evidence for medication $i$. The base model also induces an initial medication set $\hat{Y}^{(0)}_q$ under its fixed operating point. We define
\begin{equation}
    K_q = |\hat{Y}^{(0)}_q|
\end{equation}
and require every decoder to return exactly $K_q$ medications. This is the cardinality-matching condition the comparison depends on: since no decoder can alter its cardinality, none can appear to satisfy a looser DDI constraint by returning fewer drugs.

Let $G_D=(\mathcal{M},E_D)$ be an undirected DDI graph, where an edge $(i,j)\in E_D$ indicates a flagged drug--drug interaction. For any predicted set $Y$, we use the pairwise DDI rate
\begin{equation}
    \mathrm{DDI}(Y)=\frac{|\{(i,j)\in E_D:i\in Y, j\in Y\}|}{\binom{|Y|}{2}}.
\end{equation}
We use three nominal target rates $\rho\in\{0.136,0.100,0.080\}$, treated as loose, moderate, and strict prespecified DDI regimes rather than a search for an optimal operating point. Each rate is mapped to the nearest feasible integer edge budget
\begin{equation}
    B_q(\rho)=\operatorname{round}\!\left(\rho\binom{K_q}{2}\right),
\end{equation}
using the same deterministic conversion for every decoder. Because of this discretization, we report achieved DDI rates in addition to the nominal targets. The task studied here is not to learn $s(q)$; it is to solve the decoding problem: among all medication sets of size $K_q$ whose selected DDI edge count does not exceed $B_q$, choose a set that best respects the frozen scorer.

\section{Exact DDI-constrained decoding}

\subsection{MILP formulation}
For each medication $i$, introduce a binary selection variable $x_i\in\{0,1\}$. For each DDI edge $(i,j)\in E_D$, introduce an auxiliary variable $y_{ij}\in[0,1]$ indicating whether both endpoints are selected. We solve
\begin{align}
\max_{x,y}\quad & \sum_{i=1}^{M} s_i(q)x_i  \label{eq:objective}\\
\text{s.t.}\quad & \sum_{i=1}^{M} x_i = K_q, \label{eq:cardinality}\\
& \sum_{(i,j)\in E_D} y_{ij} \le B_q, \label{eq:budget}\\
& y_{ij} \ge x_i+x_j-1, \qquad \forall (i,j)\in E_D, \label{eq:linearize-lb}\\
& y_{ij} \le x_i,\quad y_{ij}\le x_j, \qquad \forall (i,j)\in E_D, \label{eq:linearize-ub}\\
& x_i\in\{0,1\},\quad 0\le y_{ij}\le 1. \label{eq:variables}
\end{align}
With binary $x_i$, constraints~\eqref{eq:linearize-lb}--\eqref{eq:linearize-ub} give the exact product $y_{ij}=x_ix_j$, so Eq.~\eqref{eq:budget} counts the selected DDI edges exactly. The formulation has 151 binary variables and 1,425 auxiliary edge variables, for 1,576 variables total (Appendix~\ref{app:impl}).

The MILP never sees the ground-truth medication set during inference; it only sees the frozen model scores, target cardinality, DDI graph, and DDI budget. At global optimality, the MILP has objective value at least as high as any heuristic solution obeying the same constraints, but a higher score sum is not guaranteed to correspond to better clinical-label agreement, nor is the same decoded set guaranteed under a rescoring of $s(q)$ that preserves rank but not magnitude. We evaluate both questions directly in Section~\ref{sec:results}.

\subsection{Why global optimization can differ from local repair}
A local repair rule compares medications inside the conflict it is currently resolving. This can be suboptimal when a high-scoring drug participates in many conflicts: retaining that drug may force the subsequent removal of several slightly lower-scoring medications whose joint score is larger. The MILP evaluates those coupled consequences simultaneously. This effect should strengthen as $B_q$ decreases, since tighter budgets reduce the number of feasible combinations and give each local commitment more downstream consequences. We evaluate whether the advantage in Jaccard grows as DDI-constraint pressure increases, and whether it depends on which baseline, operating point, or score parameterization is used.

\subsection{Implementation}
We solve the MILP with \texttt{scipy.optimize.milp}, using the HiGHS mixed-integer optimization backend \citep{virtanen2020scipy,huangfu2018highs}. No warm start, learning, or patient-specific tuning is required. Full solver settings are given in Appendix~\ref{app:impl}.

\section{Local repair baselines}

\paragraph{Greedy + refill.}
We begin from the top-$K_q$ medications under the frozen score. While the selected set exceeds the DDI budget, the decoder finds a conflicting pair and removes the lower-scoring endpoint. Once the budget is met, it scans excluded medications in descending score order and refills the set with the highest-scoring candidate whose addition preserves feasibility, continuing until cardinality $K_q$ is restored.

\paragraph{Degree-aware greedy + refill.}
This baseline uses graph structure: at each repair step, it computes every selected medication's current degree in the induced DDI subgraph and removes a medication with maximum conflict degree, breaking ties by lower model score. Refill is identical to the greedy baseline. Removing a high-degree medication can resolve several conflicts at once.

\paragraph{Degree-aware + 1-swap local search.}
Starting from the degree-aware solution, we repeatedly search every feasible 1-for-1 swap ($i\in S$ selected, $j\notin S$ excluded) and apply the swap that most increases total score while keeping the DDI budget satisfied, until no improving swap remains. Each candidate swap's resulting conflict count is evaluated in $O(1)$ via an incremental update, $C' = C - \deg_S(i) + \deg_S(j) - D_{ij}$, where $\deg_S(\cdot)$ is a medication's current conflict degree against the selected set $S$ and $C$ is the set's current conflict count, making the full swap neighborhood cheap to search at every iteration (Appendix~\ref{app:impl}). This baseline can undo a bad greedy commitment rather than only building on top of it.

\paragraph{Controlled comparison.}
All decoders receive exactly the same $s(q)$, $K_q$, DDI graph, and $B_q$, and differ only in how they search the feasible set space.

\section{Experimental setup}

\paragraph{Data and scorer.}
Patient--drug scores are taken from the frozen drug-level classifier \pcls{} released with FLAME \citep{fan2025flame}, which is not retrained or altered for this study. We use the MIMIC-III Clinical Database v1.4, a de-identified dataset distributed through PhysioNet \citep{johnson2016mimic}. The medication vocabulary contains 151 drugs. The DDI adjacency is the 151-drug graph used by FLAME's released MIMIC-III preprocessing, derived from TWOSIDES \citep{fan2025flame,tatonetti2012ddi}, and contains 1,425 undirected edges.

\paragraph{Training, validation, and test protocol.}
We use three disjoint splits of MIMIC-III visits \citep{johnson2016mimic}, following a 4:1:1 train/validation/test ratio over approximately 14,000 visits in total, with no shared patient identifiers across splits. The margin-to-cardinality scorer that defines $K_q$ (a single logistic calibration of the classifier's confidence margin) is fit exclusively on the training split, with its decision threshold selected on a held-out 20\% slice of training. This fitted scorer, together with the MILP formulation, the three local-repair baselines, and the three DDI regimes, is fixed before the validation split is examined. We report results on the validation split as a confirmatory check, then evaluate the identical, unmodified pipeline once on the test split. Test-split numbers are the primary results reported throughout; validation-split numbers are reported alongside them to show the phenomenon reproduces on an independent sample.

\paragraph{DDI regimes.}
We use target DDI rates 0.136, 0.100, and 0.080 as loose, moderate, and strict operating regimes. For the fine-grained frontier analysis (Section~\ref{sec:rq2}) we additionally sweep ten target rates from 0.14 to 0.02. We report both the nominal target and the achieved DDI rate throughout because the edge budget is integer-valued.

\paragraph{Accuracy metrics.}
For predicted set $\hat{Y}_q$ and ground-truth set $Y_q$, our primary accuracy metric is visit-level Jaccard,
\begin{equation}
J_q=\frac{|\hat{Y}_q\cap Y_q|}{|\hat{Y}_q\cup Y_q|},
\end{equation}
averaged across visits. Because all decoders output the same $K_q$ for a visit, accuracy differences cannot be attributed to different set cardinalities.

\paragraph{Statistical analysis.}
We perform 2,000 patient-cluster bootstrap resamples over \texttt{SUBJECT\_ID}, rather than resampling visits independently, and report percentile 95\% confidence intervals for the mean Jaccard difference between global decoding and each baseline.

\paragraph{Mechanism analysis.}
At the strict budget, we compare each decoded set with the patient's original pre-repair top-$K_q$ set and ground truth, recording true-positive retention and removal, false-positive removal, true-positive replacements added, and the final sum of frozen model scores.

\section{Results}
\label{sec:results}

\subsection{Global decoding improves the accuracy--DDI tradeoff}
Table~\ref{tab:main} gives the main comparison on the test split. The ordering
\begin{equation}
J_{\mathrm{global}} > J_{\mathrm{1\text{-}swap}} > J_{\mathrm{degree}} > J_{\mathrm{greedy}}
\end{equation}
holds at all three DDI regimes. At the strict target, global decoding reaches 0.4027 Jaccard, versus 0.3963 for degree-aware + 1-swap, 0.3937 for degree-aware alone, and 0.3790 for greedy. Average output size is 26.78 medications for every decoder at every budget, and the achieved DDI rate for global decoding is at or below the shared target throughout.

\begin{table}[t]
\centering
\caption{Main comparison on the test split, at matched per-patient cardinality and matched DDI budget. All methods output 26.78 medications on average.}
\label{tab:main}
\small
\begin{tabular}{llrrr}
\toprule
Target & Decoder & Actual DDI & Jaccard & Avg. size \\
\midrule
\multirow{4}{*}{0.136}
& Greedy & 0.1184 & 0.4107 & 26.78 \\
& Degree-aware & 0.1157 & 0.4155 & 26.78 \\
& Degree-aware + 1-swap & 0.1179 & 0.4186 & 26.78 \\
& \textbf{Global MILP} & 0.1176 & \textbf{0.4207} & 26.78 \\
\midrule
\multirow{4}{*}{0.100}
& Greedy & 0.0933 & 0.3920 & 26.78 \\
& Degree-aware & 0.0909 & 0.4035 & 26.78 \\
& Degree-aware + 1-swap & 0.0929 & 0.4062 & 26.78 \\
& \textbf{Global MILP} & 0.0928 & \textbf{0.4112} & 26.78 \\
\midrule
\multirow{4}{*}{0.080}
& Greedy & 0.0757 & 0.3790 & 26.78 \\
& Degree-aware & 0.0738 & 0.3937 & 26.78 \\
& Degree-aware + 1-swap & 0.0754 & 0.3963 & 26.78 \\
& \textbf{Global MILP} & 0.0754 & \textbf{0.4027} & 26.78 \\
\bottomrule
\end{tabular}
\end{table}

\subsection{The advantage is statistically distinguishable on both validation and test}
Table~\ref{tab:bootstrap} reports patient-cluster bootstrap intervals against the two strongest baselines, on both splits. Every interval excludes zero. The advantage against degree-aware + 1-swap grows from +0.0021 (loose) to +0.0064 (strict) on test, closely matching +0.0020 to +0.0061 on the independently held-out validation split; against degree-aware alone it grows from +0.0051 to +0.0090 on test and +0.0047 to +0.0089 on validation.

\begin{table}[t]
\centering
\caption{Patient-cluster bootstrap (2,000 resamples over \texttt{SUBJECT\_ID}) on the validation and test splits. All intervals exclude zero on both splits.}
\label{tab:bootstrap}
\small
\begin{tabular}{lrrrr}
\toprule
Target DDI & Global$-$1-swap & 95\% CI & Global$-$Degree & 95\% CI \\
\midrule
\multicolumn{5}{l}{\emph{Validation split}} \\
0.136 & +0.0020 & [+0.0011,+0.0030] & +0.0047 & [+0.0036,+0.0059] \\
0.100 & +0.0050 & [+0.0037,+0.0064] & +0.0083 & [+0.0068,+0.0098] \\
0.080 & +0.0061 & [+0.0047,+0.0076] & +0.0089 & [+0.0073,+0.0104] \\
\midrule
\multicolumn{5}{l}{\emph{Test split}} \\
0.136 & +0.0021 & [+0.0010,+0.0031] & +0.0051 & [+0.0040,+0.0062] \\
0.100 & +0.0050 & [+0.0036,+0.0063] & +0.0077 & [+0.0062,+0.0091] \\
0.080 & +0.0064 & [+0.0049,+0.0078] & +0.0090 & [+0.0074,+0.0106] \\
\bottomrule
\end{tabular}
\end{table}

\subsection{The advantage holds across the accuracy--DDI frontier}
\label{sec:rq2}
\begin{figure}[t]
\centering
\includegraphics[width=0.92\linewidth]{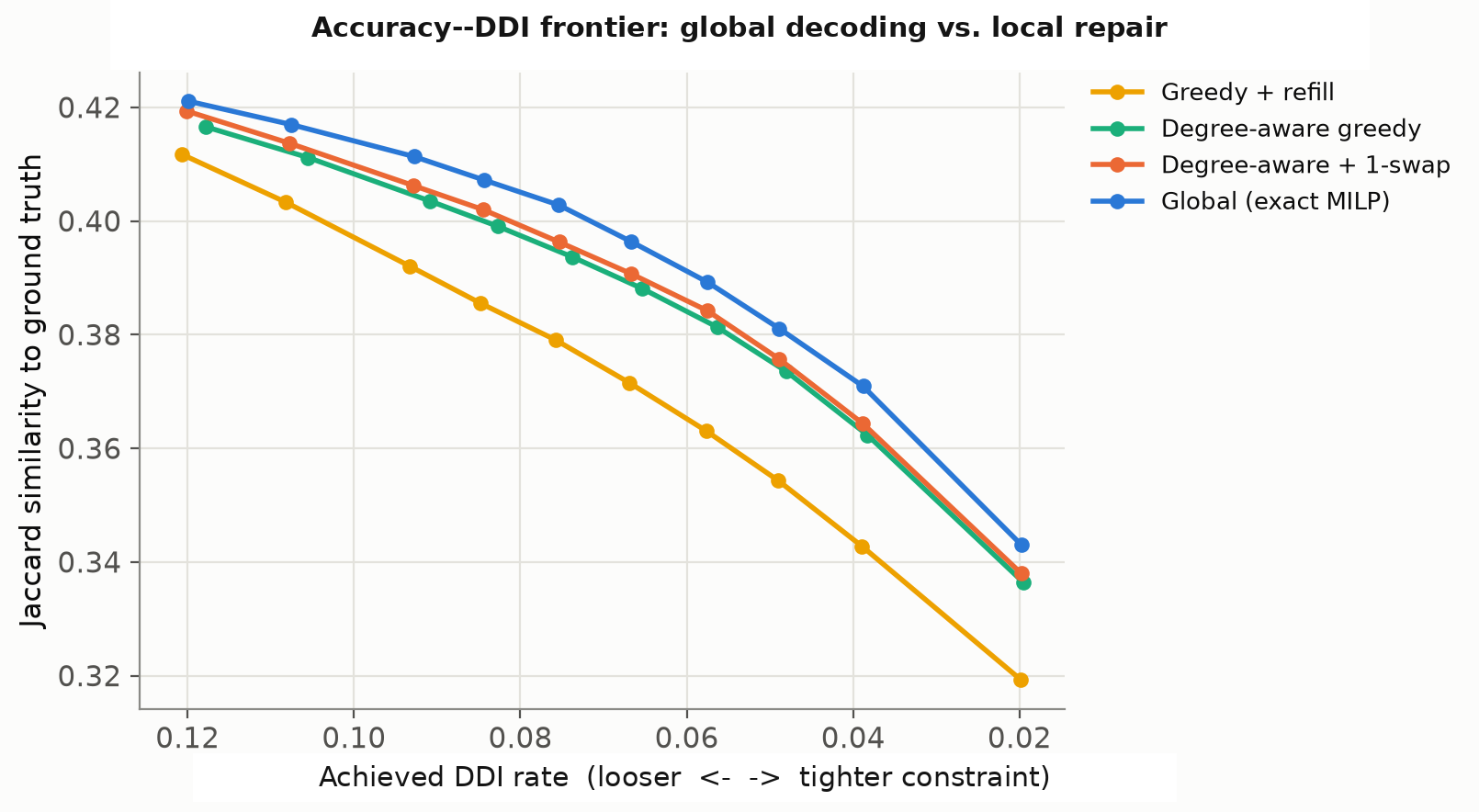}
\caption{Accuracy--DDI frontier under identical frozen scores and fixed cardinality, from a fine sweep of ten target DDI rates on the test split. The global decoder's curve remains above the three local baselines throughout, with separation widening toward stricter DDI constraints.}
\label{fig:frontier}
\end{figure}

Realized DDI rate is not always identical across decoders at a shared target, so we additionally evaluate a fine sweep of ten target rates from 0.14 to 0.02 and compare methods at matched \emph{realized} DDI values via interpolation over the overlapping range of each method's (achieved DDI, Jaccard) curve. Figure~\ref{fig:frontier} shows the resulting frontier; Table~\ref{tab:matched} reports Jaccard at matched realized-DDI checkpoints against the strongest baseline. Global decoding is above degree-aware + 1-swap at every checkpoint.

\begin{table}[t]
\centering
\caption{Jaccard at matched realized DDI rates within the overlapping range of the fine-sweep curves in Fig.~\ref{fig:frontier}.}
\label{tab:matched}
\small
\begin{tabular}{lrr}
\toprule
Realized DDI & Degree-aware + 1-swap & \textbf{Global MILP} \\
\midrule
0.12 & 0.4193 & \textbf{0.4210} \\
0.10 & 0.4098 & \textbf{0.4139} \\
0.08 & 0.3992 & \textbf{0.4050} \\
0.06 & 0.3860 & \textbf{0.3911} \\
\bottomrule
\end{tabular}
\end{table}

We additionally evaluate three score representations for the frozen scorer -- raw log-margin, sigmoid-transformed probability $P(\mathrm{Yes})$, and rank-normalization -- holding $K_q$ fixed. Greedy and degree-aware repair depend only on score order and are therefore invariant to strictly monotone rescoring. In contrast, both the 1-swap local search and the MILP use score differences or sums and can depend on score spacing. Global decoding beats degree-aware + 1-swap in all nine score-by-budget combinations tested; detailed results are reported in Appendix~\ref{app:robustness}.

The frontier is not perfectly monotonic point-to-point (Fig.~\ref{fig:frontier}), so we do not claim a strict relationship between budget tightness and gap size at every adjacent pair of points. The pattern across the coarse and fine sweeps is that global decoding provides a small advantage when the DDI constraint is permissive and a larger advantage once the constraint tightens enough that conflicting choices become coupled.

\subsection{Global decoding preserves more true positives and removes more false positives}
At the strict target, global decoding retains more true-positive medications than the local baselines while also removing more false positives, rather than trading one error type for the other. The optimized score sum is ordered in the same direction as Jaccard; the full decomposition is reported in Appendix~\ref{app:robustness}.

\section{Discussion}

\paragraph{DDI-constrained medication recommendation is also a decoding problem.}
The dominant research question in medication recommendation has been how to learn better representations and score functions. The MILP formulation itself is standard; our contribution is the controlled empirical isolation of decoder choice. Our results show that, with the scorer completely frozen, the final mapping from scores to a constrained set materially changes the accuracy--DDI frontier. A model can assign useful drug-level scores while a local DDI-repair procedure discards that information through myopic decisions.

\paragraph{Comparison with local search.}
Degree-aware repair with 1-swap local search substantially improves over greedy repair and operates at nearly the same realized DDI exposure as the global decoder at the three primary targets. Exact decoding retains an advantage of +0.0021 to +0.0064 Jaccard over this baseline across the three operating regimes on the test split, with bootstrap intervals excluding zero and the same pattern present on validation, across the fine budget sweep, and under three score representations.

\paragraph{Why the gap widens under tighter budgets.}
When the budget is permissive, many high-score sets are feasible and local commitments have limited downstream cost. When the budget is strict, the feasible region contracts and the cost of keeping one conflict-heavy drug rises. The mechanism analysis is consistent with this: globally optimized sets remove more false positives while retaining more true positives.

\paragraph{Modularity.}
The decoder assumes only patient-specific medication scores, a target cardinality, and a pairwise conflict graph, independent of how the scores were learned. Foundation models are ultimately converted from candidate-level scores into coupled medication sets, and DDI constraints can live in this inference layer as well as in model training. FLAME supplies the clinical scores; our contribution studies DDI-constrained action decoding on top of a frozen model. We instantiate one frozen scorer to keep the empirical claim narrow: decoder choice matters under controlled scores and constraints, not that this particular scorer is universally best.

\paragraph{Clinical interpretation of the DDI budget.}
The DDI graph used here is a benchmark signal, not a complete clinical decision rule, and we do not claim the resulting sets are clinically safe in an absolute sense. Our empirical claim is narrower: global decoding improves medication-set agreement at matched DDI exposure under the same TWOSIDES-derived binary interaction constraint. Interaction severity, dosage, timing, indication, allergy history, organ function, and outcome data are not represented in this constraint, and clinician oversight remains necessary. The three budgets are controlled operating regimes for studying algorithmic behavior, not prescriptions for a clinically acceptable DDI rate.

\section{Limitations}
Our claims are intentionally scoped to one EHR benchmark, one frozen scorer, and a binary TWOSIDES-derived DDI graph \citep{tatonetti2012ddi}. The fixed-cardinality design isolates decoder choice but does not optimize regimen size, and the MILP inherits any systematic bias in the underlying scores. The DDI constraint is a benchmark safety signal rather than a complete clinical rule because it omits interaction severity and patient-specific context. Accordingly, the present evaluation does not establish the clinical safety or clinical utility of the resulting recommended medication sets. Broader validation across datasets, scorers, patient populations, and richer clinically informed constraints would be needed before evaluating the approach for potential clinical deployment. Runtime and scaling under larger vocabularies or richer constraints, as well as comparisons with stronger multi-swap or beam-search baselines, remain important directions for future evaluation.

\section{Conclusion}
Medication recommendation is commonly treated as a representation-learning problem followed by a relatively simple set-construction step. We isolate that final step and show that it matters. With the scorer, output cardinality, and DDI budget held fixed, in this benchmark evaluation, exact global decoding achieves higher Jaccard than increasingly sophisticated local repair on validation and test, at matched realized DDI exposure and across multiple score parameterizations. These findings are limited to benchmark medication recommendations under the evaluated DDI constraint and do not establish clinical safety or utility; they do not prescribe how medication decisions should generally be made. Within this evaluation, when DDI constraints couple medication choices, the final set can be treated as a joint constrained decision rather than a sequence of local repairs.

\clearpage
\bibliographystyle{plainnat}
\bibliography{references}

@article{ali2023survey,
  author = {Ali, Zafar and Huang, Yi and Ullah, Irfan and Feng, Junlan and Deng, Chao and Thierry, Nimbeshaho and Khan, Asad and Jan, Asim Ullah and Shen, Xiaoli and Wu, Rui and Qi, Guilin},
  title = {Deep learning for medication recommendation: A systematic survey},
  journal = {Data Intelligence},
  volume = {5},
  number = {2},
  pages = {303--354},
  year = {2023},
  doi = {10.1162/dint_a_00197}
}

@inproceedings{chen2023carmen,
  author = {Chen, Qianyu and Li, Xin and Geng, Kunnan and Wang, Mingzhong},
  title = {Context-aware safe medication recommendations with molecular graph and DDI graph embedding},
  booktitle = {Proceedings of the AAAI Conference on Artificial Intelligence},
  volume = {37},
  pages = {7053--7060},
  year = {2023},
  doi = {10.1609/aaai.v37i6.25861}
}

@inproceedings{fan2025flame,
  author = {Fan, Chenxiao and Gao, Chongming and Shi, Wentao and Gong, Yaxin and Zhao, Zihao and Feng, Fuli},
  title = {Fine-grained list-wise alignment for generative medication recommendation},
  booktitle = {Advances in Neural Information Processing Systems},
  volume = {38},
  pages = {53569--53593},
  year = {2025},
  doi = {10.52202/085713-1602}
}

@article{huangfu2018highs,
  author = {Huangfu, Qi and Hall, J. A. Julian},
  title = {Parallelizing the dual revised simplex method},
  journal = {Mathematical Programming Computation},
  volume = {10},
  number = {1},
  pages = {119--142},
  year = {2018},
  doi = {10.1007/s12532-017-0130-5}
}

@article{huh2026pace,
  author = {Huh, Chaeyoung and Hwang, Hyunmin and Shin, Jung Hwan and Park, Jinse and Ye, Jong Chul},
  title = {PACE-RAG: Patient-aware contextual and evidence-based policy RAG for clinical drug recommendation},
  journal = {arXiv preprint arXiv:2603.17356},
  year = {2026}
}

@article{johnson2016mimic,
  author = {Johnson, Alistair E. W. and Pollard, Tom J. and Shen, Lu and Lehman, Li-wei H. and Feng, Mengling and Ghassemi, Mohammad and Moody, Benjamin and Szolovits, Peter and Celi, Leo Anthony and Mark, Roger G.},
  title = {MIMIC-III, a freely accessible critical care database},
  journal = {Scientific Data},
  volume = {3},
  pages = {160035},
  year = {2016},
  doi = {10.1038/sdata.2016.35}
}

@inproceedings{kim2024vita,
  author = {Kim, Taeri and Heo, Jiho and Kim, Hongil and Shin, Kijung and Kim, Sang-Wook},
  title = {VITA: ``Carefully chosen and weighted less'' is better in medication recommendation},
  booktitle = {Proceedings of the AAAI Conference on Artificial Intelligence},
  volume = {38},
  pages = {8600--8607},
  year = {2024},
  doi = {10.1609/aaai.v38i8.28704}
}

@inproceedings{le2018dmnc,
  author = {Le, Hung and Tran, Truyen and Venkatesh, Svetha},
  title = {Dual memory neural computer for asynchronous two-view sequential learning},
  booktitle = {Proceedings of the 24th ACM SIGKDD International Conference on Knowledge Discovery and Data Mining},
  pages = {1637--1645},
  year = {2018},
  doi = {10.1145/3219819.3219981}
}

@inproceedings{li2024causalmed,
  author = {Li, Xiang and Liang, Shunpan and Lei, Yu and Li, Chen and Hou, Yulei and Zheng, Dashun and Ma, Tengfei},
  title = {CausalMed: Causality-based personalized medication recommendation centered on patient health state},
  booktitle = {Proceedings of the 33rd ACM International Conference on Information and Knowledge Management},
  pages = {1276--1285},
  year = {2024},
  doi = {10.1145/3627673.3679542}
}

@inproceedings{li2025timeaware,
  author = {Li, Yishuo and Zhang, Qi and Lu, Wenpeng and Peng, Xueping and Zhang, Weiyu and Si, Jiasheng and Gong, Yongshun and Hu, Liang},
  title = {Time-aware medication recommendation via intervention of dynamic treatment regimes},
  booktitle = {Proceedings of the ACM on Web Conference 2025},
  pages = {5161--5172},
  year = {2025},
  doi = {10.1145/3696410.3714533}
}

@article{liu2024leader,
  author = {Liu, Qidong and Wu, Xian and Zhao, Xiangyu and Zhu, Yuanshao and Zhang, Zijian and Tian, Feng and Zheng, Yefeng},
  title = {Large language model distilling medication recommendation model},
  journal = {arXiv preprint arXiv:2402.02803},
  year = {2024}
}

@inproceedings{lv2026giant,
  author = {Lv, Hang and Guo, Zixuan and Tan, Yanchao and Shao, Wanzi and Zhang, Hengyu and Yang, Carl},
  title = {Awaken the giant: Activating LLMs via deep model guidance for boundary-aware medication recommendation},
  booktitle = {Proceedings of the 32nd ACM SIGKDD Conference on Knowledge Discovery and Data Mining},
  pages = {1030--1041},
  year = {2026},
  doi = {10.1145/3770854.3780297}
}

@book{nemhauser1988integer,
  author = {Nemhauser, George L. and Wolsey, Laurence A.},
  title = {Integer and Combinatorial Optimization},
  publisher = {Wiley},
  year = {1988}
}

@inproceedings{shang2019gamenet,
  author = {Shang, Junyuan and Xiao, Cao and Ma, Tengfei and Li, Hongyan and Sun, Jimeng},
  title = {GAMENet: Graph augmented memory networks for recommending medication combination},
  booktitle = {Proceedings of the AAAI Conference on Artificial Intelligence},
  volume = {33},
  pages = {1126--1133},
  year = {2019},
  doi = {10.1609/aaai.v33i01.33011126}
}

@inproceedings{shang2019gbert,
  author = {Shang, Junyuan and Ma, Tengfei and Xiao, Cao and Sun, Jimeng},
  title = {Pre-training of graph augmented transformers for medication recommendation},
  booktitle = {Proceedings of the Twenty-Eighth International Joint Conference on Artificial Intelligence},
  pages = {5953--5959},
  year = {2019},
  doi = {10.24963/ijcai.2019/825}
}

@inproceedings{tan2022foursdrug,
  author = {Tan, Yanchao and Kong, Chengjun and Yu, Leisheng and Li, Pan and Chen, Chaochao and Zheng, Xiaolin and Hertzberg, Vicki S. and Yang, Carl},
  title = {4SDrug: Symptom-based set-to-set small and safe drug recommendation},
  booktitle = {Proceedings of the 28th ACM SIGKDD Conference on Knowledge Discovery and Data Mining},
  pages = {3970--3980},
  year = {2022},
  doi = {10.1145/3534678.3539089}
}

@inproceedings{tan2024nla,
  author = {Tan, Jie and Rong, Yu and Zhao, Kangfei and Bian, Tian and Xu, Tingyang and Huang, Junzhou and Cheng, Hong and Meng, Helen},
  title = {Natural language-assisted multi-modal medication recommendation},
  booktitle = {Proceedings of the 33rd ACM International Conference on Information and Knowledge Management},
  pages = {2200--2209},
  year = {2024},
  doi = {10.1145/3627673.3679529}
}

@article{tatonetti2012ddi,
  author = {Tatonetti, Nicholas P. and Ye, Patrick P. and Daneshjou, Roxana and Altman, Russ B.},
  title = {Data-driven prediction of drug effects and interactions},
  journal = {Science Translational Medicine},
  volume = {4},
  number = {125},
  pages = {125ra31},
  year = {2012},
  doi = {10.1126/scitranslmed.3003377}
}

@article{virtanen2020scipy,
  author = {Virtanen, Pauli and Gommers, Ralf and Oliphant, Travis E. and Haberland, Matt and Reddy, Tyler and others},
  title = {SciPy 1.0: Fundamental algorithms for scientific computing in Python},
  journal = {Nature Methods},
  volume = {17},
  pages = {261--272},
  year = {2020},
  doi = {10.1038/s41592-019-0686-2}
}

@inproceedings{wu2022cognet,
  author = {Wu, Rui and Qiu, Zhaopeng and Jiang, Jiacheng and Qi, Guilin and Wu, Xian},
  title = {Conditional generation net for medication recommendation},
  booktitle = {Proceedings of the ACM Web Conference 2022},
  pages = {935--945},
  year = {2022},
  doi = {10.1145/3485447.3511936}
}

@article{wu2023dapsnet,
  author = {Wu, Jialun and Dong, Yuxin and Gao, Zeyu and Gong, Tieliang and Li, Chen},
  title = {Dual attention and patient similarity network for drug recommendation},
  journal = {Bioinformatics},
  volume = {39},
  number = {1},
  pages = {btad003},
  year = {2023},
  doi = {10.1093/bioinformatics/btad003}
}

@inproceedings{wu2025armr,
  author = {Wu, Feiyue and Wu, Tianxing and Jing, Shenqi},
  title = {ARMR: Adaptively responsive network for medication recommendation},
  booktitle = {Proceedings of the Thirty-Fourth International Joint Conference on Artificial Intelligence},
  pages = {7831--7839},
  year = {2025},
  doi = {10.24963/ijcai.2025/871}
}

@inproceedings{yang2021micron,
  author = {Yang, Chaoqi and Xiao, Cao and Glass, Lucas and Sun, Jimeng},
  title = {Change matters: Medication change prediction with recurrent residual networks},
  booktitle = {Proceedings of the Thirtieth International Joint Conference on Artificial Intelligence},
  pages = {3728--3734},
  year = {2021},
  doi = {10.24963/ijcai.2021/513}
}

@inproceedings{yang2021safedrug,
  author = {Yang, Chaoqi and Xiao, Cao and Ma, Fenglong and Glass, Lucas and Sun, Jimeng},
  title = {SafeDrug: Dual molecular graph encoders for recommending effective and safe drug combinations},
  booktitle = {Proceedings of the Thirtieth International Joint Conference on Artificial Intelligence},
  pages = {3735--3741},
  year = {2021},
  doi = {10.24963/ijcai.2021/514}
}

@inproceedings{yang2023molerec,
  author = {Yang, Nianzu and Zeng, Kaipeng and Wu, Qitian and Yan, Junchi},
  title = {MoleRec: Combinatorial drug recommendation with substructure-aware molecular representation learning},
  booktitle = {Proceedings of the ACM Web Conference 2023},
  pages = {4075--4085},
  year = {2023},
  doi = {10.1145/3543507.3583872}
}

@inproceedings{zhang2017leap,
  author = {Zhang, Yutao and Chen, Robert and Tang, Jie and Stewart, Walter F. and Sun, Jimeng},
  title = {LEAP: Learning to prescribe effective and safe treatment combinations for multimorbidity},
  booktitle = {Proceedings of the 23rd ACM SIGKDD International Conference on Knowledge Discovery and Data Mining},
  pages = {1315--1324},
  year = {2017},
  doi = {10.1145/3097983.3098109}
}

@inproceedings{zhao2024raremed,
  author = {Zhao, Zihao and Jing, Yi and Feng, Fuli and Wu, Jiancan and Gao, Chongming and He, Xiangnan},
  title = {Leave no patient behind: Enhancing medication recommendation for rare disease patients},
  booktitle = {Proceedings of the 47th International ACM SIGIR Conference on Research and Development in Information Retrieval},
  pages = {533--542},
  year = {2024},
  doi = {10.1145/3626772.3657785}
}

@article{zhao2025lamo,
  author = {Zhao, Zihao and Fan, Chenxiao and Gao, Chongming and Feng, Fuli and He, Xiangnan},
  title = {Addressing overprescribing challenges: Fine-tuning large language models for medication recommendation tasks},
  journal = {arXiv preprint arXiv:2503.03687},
  year = {2025}
}

\clearpage
\appendix
\section{Implementation details}
\label{app:impl}

\paragraph{Data splits.} The training, validation, and test splits (4:1:1 ratio, approximately 14,000 visits in total) share no \texttt{SUBJECT\_ID} across splits. The medication vocabulary ($M=151$) and DDI graph ($|E_D|=1{,}425$) are shared across all splits and taken from FLAME's released MIMIC-III preprocessing \citep{fan2025flame}. We do not use FLAME's optional clinical-note augmentation step (\texttt{data4LLM\_with\_note.csv}, which requires GPT API access to extract note-derived features); our scores are computed from structured MIMIC-III fields only.

\paragraph{Scorer fitting.} The cardinality-defining scorer is a one-dimensional logistic regression mapping the frozen classifier's confidence margin to a calibrated probability, fit by maximum likelihood on an 80\% slice of the training split (\texttt{scikit-learn}, \texttt{LogisticRegression}, $L_2$ penalty, $C=1$, \texttt{max\_iter}=1000). The decision threshold is selected on the remaining 20\% slice of training by maximizing mean Jaccard over a grid $\{0.05, 0.07, \ldots, 0.93\}$. The selected threshold (0.310) and fitted coefficients are frozen and applied unchanged to the validation and test splits.

\paragraph{MILP solver settings.} We use \texttt{scipy.optimize.milp} with the HiGHS backend \citep{virtanen2020scipy,huangfu2018highs} and default branching and cutting-plane settings. Every reported instance reached proven global optimality. Edge variables $y_{ij}$ are declared continuous on $[0,1]$ rather than binary, since the linearization constraints~\eqref{eq:linearize-lb}--\eqref{eq:linearize-ub} force them to $\{0,1\}$ given binary $x$; this substantially reduces the number of integer variables relative to declaring $y_{ij}$ binary, with no change to the optimal solution.

\paragraph{1-swap local search.} Given a starting feasible set $S$ (the degree-aware solution) and a DDI budget $B_q$, we maintain the per-medication conflict degree against the current set, $\deg_S(d) = \sum_{s \in S} D_{ds}$, as a length-151 vector, updated incrementally after each swap rather than recomputed from scratch. At each iteration we evaluate, for every pair $(i \in S, j \notin S)$, the resulting conflict count $C' = C - \deg_S(i) + \deg_S(j) - D_{ij}$ and score change $s_j - s_i$, select the feasible swap ($C' \le B_q$) with the largest score improvement, apply it, and update $\deg_S$. The loop terminates when no feasible improving swap exists or after 200 iterations (never reached in practice).

\paragraph{Score representations (Section~\ref{sec:rq2}).} Raw log-margin is the classifier's native output $z_i(q)$. Sigmoid probability is $P(\mathrm{Yes}\mid q, m_i) = 1/(1+e^{-z_i(q)})$. Rank-normalization replaces $z_i(q)$ with $(150-\mathrm{rank}_i)/150$, where $\mathrm{rank}_i\in\{0,\ldots,150\}$ is the 0-indexed descending rank of drug $i$ under $z(q)$ for that visit; this preserves the descending medication order while changing the relative spacing between scores. Greedy and degree-aware repair depend only on this order, whereas both 1-swap local search and the MILP can depend on the score spacing.

\clearpage
\section{Additional robustness and mechanism details}
\label{app:robustness}
The main text summarizes two secondary analyses that support the central decoding result. Table~\ref{tab:scoresens} reports the score-representation sensitivity analysis, and Table~\ref{tab:mechanism} gives the strict-budget error decomposition.

\begin{table}[h!]
\centering
\caption{Global decoding vs.\ degree-aware + 1-swap under three score representations, at each DDI regime.}
\label{tab:scoresens}
\small
\begin{tabular}{lccc}
\toprule
Score representation & Loose (0.136) & Moderate (0.100) & Strict (0.080) \\
\midrule
Raw log-margin & Global\,\cmark & Global\,\cmark & Global\,\cmark \\
Sigmoid $P(\mathrm{Yes})$ & Global\,\cmark & Global\,\cmark & Global\,\cmark \\
Rank-normalized & Global\,\cmark & Global\,\cmark & Global\,\cmark \\
\bottomrule
\end{tabular}
\end{table}

\begin{table}[h!]
\centering
\caption{Mechanism analysis at target DDI 0.08, relative to the original pre-repair top-$K_q$ set. Values are per-patient averages.}
\label{tab:mechanism}
\small
\begin{tabular}{lrrrrr}
\toprule
Decoder & TP retained & TP removed & FP removed & TP repl. & Score sum \\
\midrule
Greedy & 13.175 & 2.042 & 1.829 & 0.810 & 14.6425 \\
Degree-aware & 13.717 & 1.500 & 1.385 & 0.693 & 15.0374 \\
Degree-aware + 1-swap & 13.790 & 1.427 & 1.458 & 0.693 & 15.1224 \\
\textbf{Global MILP} & \textbf{13.858} & \textbf{1.359} & \textbf{2.049} & 0.799 & \textbf{15.3485} \\
\bottomrule
\end{tabular}
\end{table}

\clearpage
\section{Additional bootstrap visualization}
\label{app:bootstrapviz}
\begin{figure}[h!]
\centering
\includegraphics[width=0.86\linewidth]{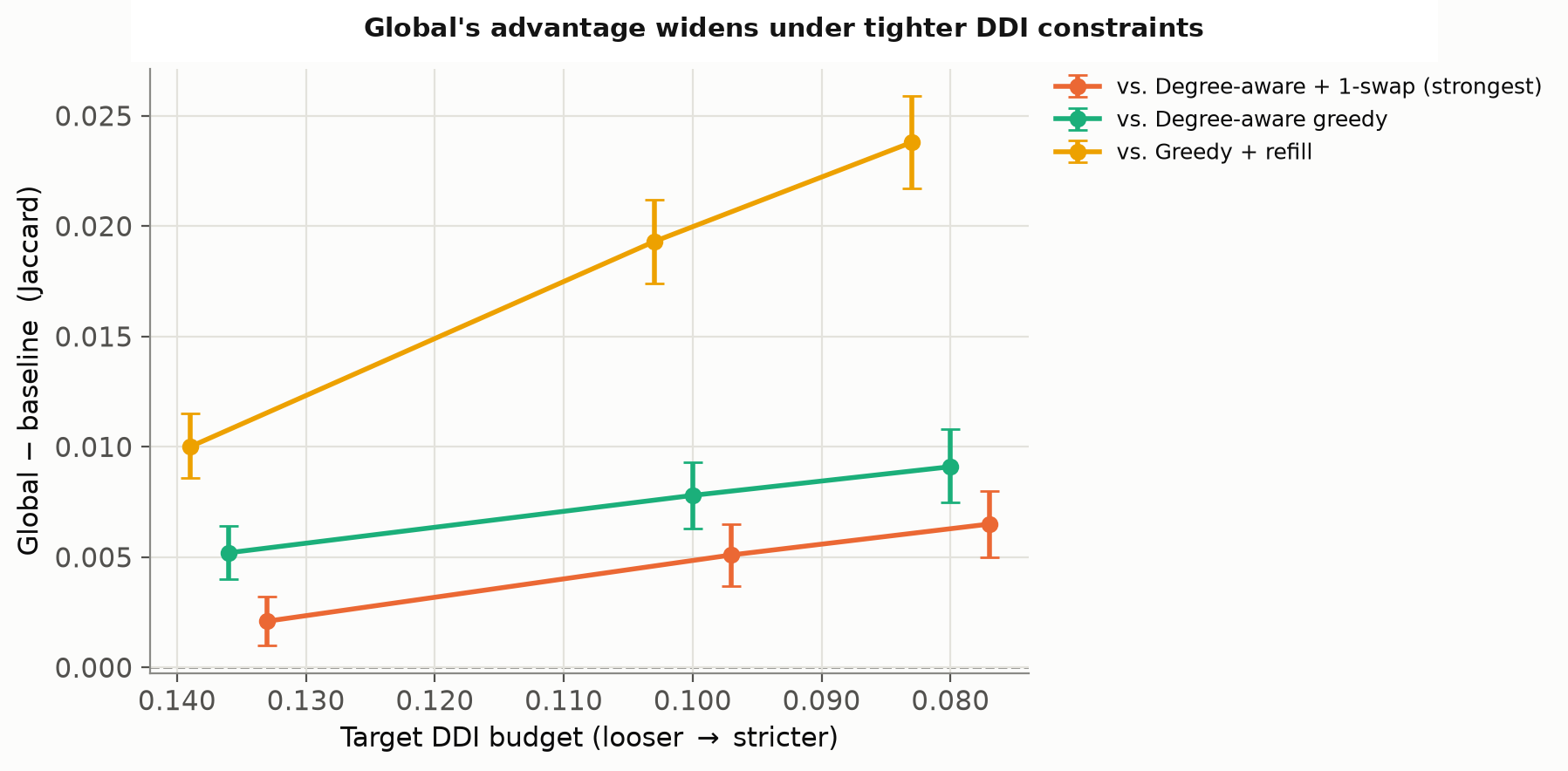}
\caption{Global decoding's advantage over each baseline as the DDI budget tightens, with patient-cluster bootstrap 95\% confidence intervals on the test split.}
\label{fig:advantage}
\end{figure}

Figure~\ref{fig:advantage} summarizes the patient-cluster bootstrap advantage over each local baseline on the test split.

\end{document}